\documentclass[runningheads]{llncs}
\usepackage[pagebackref,breaklinks,colorlinks]{hyperref}
\usepackage{graphicx}
\usepackage[small]{caption}
\usepackage{graphicx}
\usepackage{amsmath}
\usepackage{amssymb}
\usepackage{booktabs}
\usepackage{algorithm}
\usepackage{algorithmic}
\usepackage[switch]{lineno}

\usepackage{color,wrapfig}
\usepackage{tabularx}
\usepackage{multirow}
\usepackage{booktabs}
\usepackage{makecell}
\usepackage{enumitem}
\setitemize{noitemsep,topsep=0pt,parsep=0pt,partopsep=0pt}
\usepackage{caption}
\usepackage{wrapfig}
\usepackage[capitalize]{cleveref}
\crefname{section}{Sec.}{Secs.}
\Crefname{section}{Section}{Sections}
\Crefname{table}{Table}{Tables}
\crefname{table}{Tab.}{Tabs.}

\begin{document}
\title{Rethinking Timesteps Samplers \\ and Prediction Types}

% \author{Paper ID: 1669}
% \institute{}
%
\titlerunning{Rethinking Timesteps Samplers and Prediction Types}

\author{Bin Xie\inst{1} \and
Gady Agam\inst{1}}

% % TODO FINAL: Replace with an abbreviated list of authors.
\authorrunning{B.~Xie et al.}

\institute{Department of Computer Science, Illinois Institute of Technology, USA \\ 
\email{\small{bxie9@hawk.iit.edu, agam@iit.edu}}
}

\maketitle              % typeset the header of the contribution
%
% \vspace{-0.4cm}
\begin{abstract}
Diffusion models suffer from the huge consumption of time and resources to train. For example, diffusion models need hundreds of GPUs to train for several weeks for a high-resolution generative task to meet the requirements of an extremely large number of iterations and a large batch size. Training diffusion models become a millionaire's game. With limited resources that only fit a small batch size, training a diffusion model always fails. 
In this paper, we investigate the key reasons behind the difficulties of training diffusion models with limited resources. Through numerous experiments and demonstrations, we identified a major factor: the significant variation in the training losses across different timesteps, which can easily disrupt the progress made in previous iterations.
Moreover, different prediction types of $x_0$ exhibit varying effectiveness depending on the task and timestep. We hypothesize that using a mixed-prediction approach to identify the most accurate $x_0$ prediction type could potentially serve as a breakthrough in addressing this issue. In this paper, we outline several challenges and insights, with the hope of inspiring further research aimed at tackling the limitations of training diffusion models with constrained resources, particularly for high-resolution tasks.


\end{abstract}
\section{Introduction and Motivation}
Recent works on diffusion-based~(DDPM~\cite{ho2020denoising}) or scored-based~(NCSN~\cite{song2020score}) generative models which we call both models “diffusion models” for brevity, have proposed with similar ideas underneath but two kinds of different perspectives. Firstly, the \textit{diffusion process} utilizes $T$ steps of a small amount of isotropic Gaussian noise with gradually incremental standard deviations to corrupt a data $x_0 \sim q(x_0)$. Eventually when $T$ is sufficiently large $T\rightarrow\infty$, $x_{T}$ is equivalent to an isotropic Gaussian distribution. Then diffusion models are trained to learn to denoise each different step. Finally, diffusion models can construct desired data samples via a Markov chain that progressively denoises from a Gaussian noise into a high-quality image. The Markov process of diffusion models is either based on Langevin dynamics algorithm~\cite{song2020score} or learned via reversing the above \textit{diffusion process} for score-based or diffusion-based generative models\cite{sohl2015deep}, respectively.

Both breakthrough diffusion models of DDPM~\cite{ho2020denoising} and NCSN~\cite{song2020score} have achieved excellent performance on unconditional image generation tasks. Afterward, several recent papers have proposed to improve the efficiencies and effectiveness of diffusion models, such as DDIM~\cite{song2020denoising} improved the speed of the inference process, improved DDPM~\cite{nichol2021improved} improves the noise schedule and replaces the fixing $\sigma_{t}^{2}$ to a learnable $\Sigma_{\theta}(x_{t}, t)$ to improve the performance, and Guided-Diffusion~\cite{dhariwal2021diffusion} achieves state-of-the-art performance via modifying the network architecture of the UNet~\cite{ronneberger2015u} and involving a guided classifier to achieve conditional diffusion models on class labels. 

However, diffusion models suffer from the huge consumption of time and resources to train. For example, training diffusion models for high-resolution generative tasks demands hundreds of GPUs and several weeks of processing to accommodate the vast number of iterations and large batch sizes. Specifically, training a Stable Diffusion~\cite{rombach2022high} model utilized 256 A100 GPUs and spent 32 days, of which the total cost is nearly half a million dollars. Training diffusion models become a millionaire's game. Unfortunately, when resources are limited to small batch sizes, training a diffusion model often ends in failure. Why is it so difficult to train a diffusion model? Is it possible to train one with limited resources? In this paper, we explore the reasons why training diffusion models with constrained resources is such a formidable challenge.

The key reason why diffusion models need so many resources to train is ideally to expect to train a neural network via involving timesteps $t$ to approximate $T$~(such as 1000) number of neural networks to handle and predict $\boldsymbol{\epsilon}_t$. We found that there exist differences in MSE losses between the prediction of noise $\boldsymbol{\epsilon}_\theta$ and the real noise $\boldsymbol{\epsilon}_t$ by a large order of magnitude along with timesteps. The difference between the maximum loss~($t=0$) and the minimum loss~($t=T$) is usually on the order of $10^6$ magnitude. Under such extremely harsh demands, training diffusion models are compelled to utilize so many resources. 

Meanwhile, the sampling of diffusion models during the training process is a random sampling strategy from the range of $[0, T]$. When we sample a small batch of timesteps and finish the back-propagation, how to promise the loss of the total timesteps decrease together? 

However, training diffusion models still fail when we increase the number of iterations with a small batch size so that the total number of sampling for each timestep is equal to the number of sampling for each timestep of the diffusion models with a large batch size. The same resources would not result in the same results. 

Through numerous experiments and demonstrations, we discovered significant differences in sampling timesteps when using small batch sizes. These differences lead to large variations in MSE losses during training. Specifically, an iteration with a large MSE loss will destroy the previous iterations with a relatively smaller MSE loss if the losses are not of the same magnitude. As a result, the large discrepancies in sampling timesteps often undermine the progress from earlier efforts. 
Additionally, we found that different timesteps have varying levels of impact, with timesteps closer to 0 playing a more crucial role during the inference process. Therefore, when working with limited resources, it is essential to allocate more resources to the more important timesteps.
% Therefore, we should include the timesteps of different-scale MSE losses as much as possible in one iteration. In this way, different-scale MSE losses will be decreased throughout the whole training process.

% training diffusion models near to zero will destroy the effort of training the timesteps near to T if we randomly sample timesteps with a small batch size.

% Meanwhile, we found different timesteps contribute to different effects. Timesteps closer to 0 contribute more during the inference process. Therefore, we should allocate more resources in the more important timesteps with limited resources.

Moreover, there are three parameterizations to predict $\mathbf{x}_0$ from the input $\mathbf{x}_t$ at time step $t$, predicting $x_0$ directly, $v_t$, and $\epsilon_t$. Different types of predictions have different abilities in terms of different tasks and different timesteps. In order to mitigate the variant, we equip our model with the three types of predictions so that our model can achieve more accurate predictions of $x_0$ via only doing gradient descent for the minimum loss of the three predictions during the training process.  
% we propose a Mixed Predictions to handle different tasks.

To summarize, we find the following issues: i) we found the key issue for training diffusion models with small batch sizes is that different-scale training losses from different timesteps always destroy each other via a uniform timesteps sampler. 
% Therefore, we propose a timesteps sampler in that we first calculate the MSE losses for each timestep from pre-trained models, and then we split all timesteps $T$ into the slots with the same batch size. For the range of each slot, the sum of MSE losses for all timesteps in each slot is the same from $t=0$ to $t=T$. 
ii) We hypothesize that using the mixed-prediction of $x_0$ via the equipment of three prediction types mitigates the variant of the abilities in terms of different tasks and different timesteps. 

\section{Observations and Rethinking}
\begin{figure}[ht!]
\centering
\includegraphics[width=1.0\textwidth]{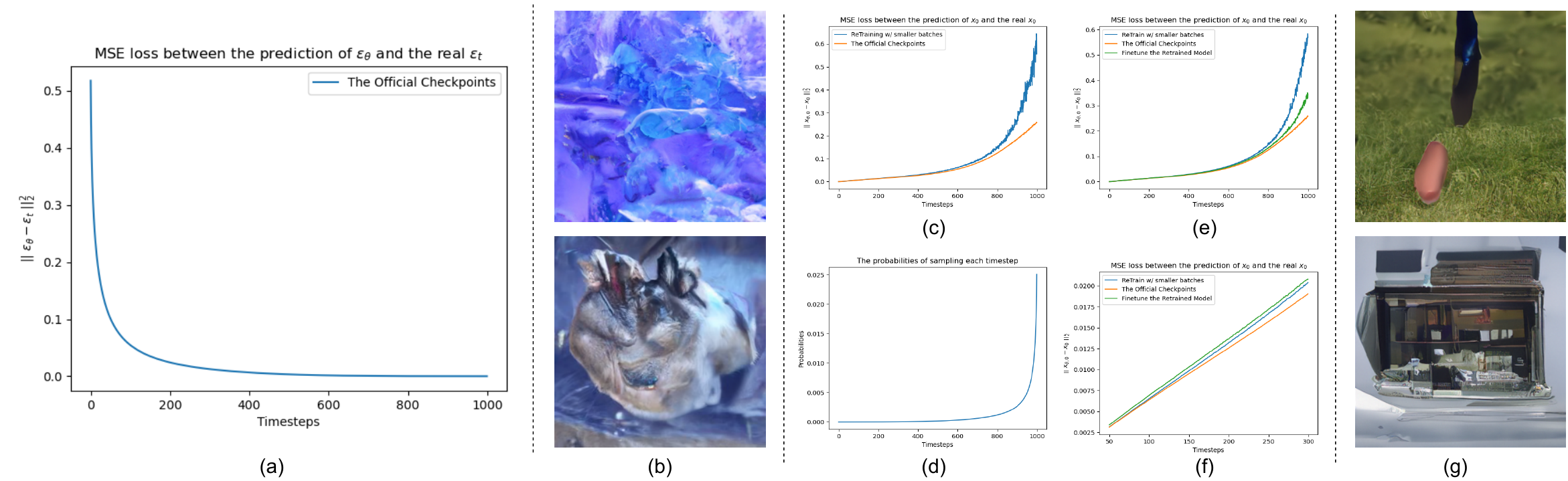}
      \caption{The timesteps sampler.} 
% \textcolor{red}{[include more detailed descriptions.]}}
  \vspace{-0.4cm}
\label{fig:tsampler}
\end{figure}
In this section, we describe the settings and observations of diffusion models, based on Guided-Diffusion~\cite{dhariwal2021diffusion} experimental settings, to exploit the intrinsic attributes about the sampling of the timesteps and types of the prediction of $\mathbf{x}_0$. 

\subsection{Rethink Timesteps Sampler}
Although diffusion models have strong abilities in image synthesis tasks, the consumption of resources and time is severely huge. It usually spends hundreds of GPUs for more than one month to train a high-resolution diffusion model, for example, training a Stable Diffusion model utilized 256 A100 GPUs and spent 32 days, of which the total cost is nearly half a million dollars. The total number of training iterations usually is more than 1 million. The reason why diffusion models need so many training iterations is ideally to expect to train a neural network via involving timesteps $t$ to approximate $T$~(such as 1000) number of neural networks to handle and predict $\boldsymbol{\epsilon}_t$. For each timestep of the neural network, it needs enough iterations to train a good model. 

But when we utilize the same ImageNet256 generative diffusion model from Guided-Diffusion~\cite{dhariwal2021diffusion} with a smaller batch size and relatively more training iterations to achieve the same training iterations for each timestep, the final model achieves a poor performance. Figure~\ref{fig:tsampler} (b) shows some generative results, which are poor performance. We propose a question: Why do not enough training iterations for each timestep produce a good model with small batch sizes? 

Based on the above bad model, we calculate the MSE loss between the prediction of $x_0$ with the real $x_0$ for each timestep. Figure~\ref{fig:tsampler} (c) illustrate the results. As we can see, the main difference comes from the timesteps nearly $t=T$. Based on the observations, we retrain the bad model with the timesteps sampler the probabilities of each timestep are shown in Figure~\ref{fig:tsampler} (d). In this way, we let the bad model train more iterations for the timesteps nearly $t=T$ so that it can decrease the difference between the pre-trained model and the bad model. After 100k training iterations, the difference does decrease shown in Figure~\ref{fig:tsampler} (e). However, the losses of the timesteps closed to $t=0$ instead increase shown in Figure~\ref{fig:tsampler} (f). The retrained model~(the green line) has larger losses compared with the previous bad mode. It demonstrates training the timesteps close to $t=T$ cause the timesteps close to $t=0$ to be worse. 
Meanwhile, the generative results are also poor performance shown in Figure~\ref{fig:tsampler} (g). One main conjecture is that the performance of the model at the timesteps nearly $t=0$ has been destroyed when many iterations at the timesteps nearly $t=T$ are processed. 

Meanwhile, we utilize the pre-trained ImageNet256 generative diffusion model to calculate the MSE loss between the prediction of noise $\boldsymbol{\epsilon}_\theta$ and the real noise $\boldsymbol{\epsilon}_t$ for each timestep. The results are shown in Figure~\ref{fig:tsampler} (a). As we can see, there are severe differences from $t=0$ to $t=T$. The difference between the maximum loss and the minimum loss is usually on the order of $10^6$ magnitude. It is extremely hard to converge due to the severe differences in the training process. 

Therefore, based on the above analysis, we split the whole timesteps into ten slots based on the MSE losses between the prediction of noise $\boldsymbol{\epsilon}_\theta$ and the real noise $\boldsymbol{\epsilon}_t$ for each timestep. The whole procedure is that we first sum the MSE loss values of all timesteps, and then split all timesteps into each slot so that the sum of MSE losses for all timesteps in each slot is the same from $t=0$ to $t=T$. The specific slots show in Table~\ref{tab:slots}. We selected the 3rd, 5th, 7th, and 9th slots to finetune. 

\begin{table}[htbp!]
    \centering
    \resizebox{0.99\textwidth}{!}{ %< auto-adjusts font size to fill line
    \begin{tabular}{@{}l|c|c|c|c|c|c|c|c|c|c@{}}
    \toprule
    & 1st & 2nd  & 3rd & 4th & 5th & 6th  & 7th & 8th  & 9th & 10th \\ \midrule
    Range & [0,4] & [5, 12] & [13,22] & [23,36] & [37,55] & [56,81] & [82,119] & [120,176] & [177,276] & [277, 1000] \\
    \bottomrule
    \end{tabular}}
    \caption{
    The slots.
    } % \caption
    \label{tab:slots}
    %\vspace{-0.8cm}
\end{table}

\begin{figure}[t]
\centering
\includegraphics[width=1.0\textwidth]{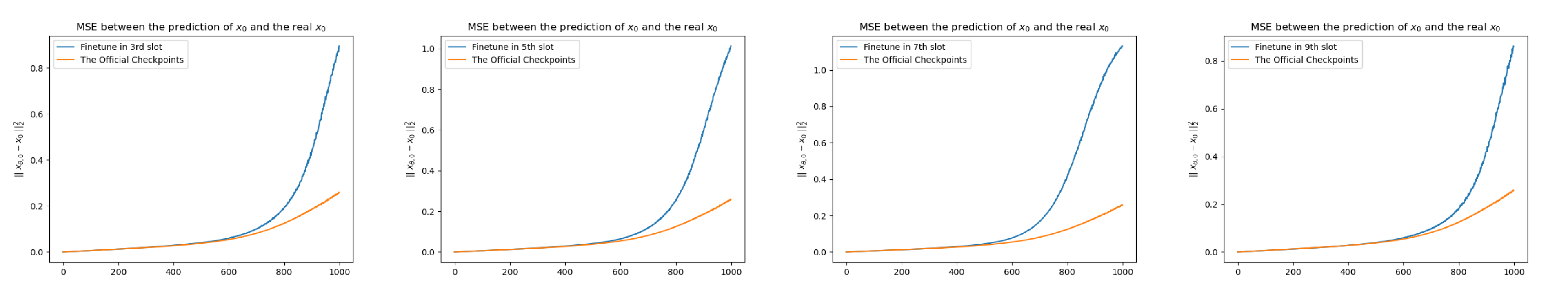}
      \caption{The results after training many iterations for the 3rd, 5th, 7th, and 9th slots of timesteps.} 
% \textcolor{red}{[include more detailed descriptions.]}}
  \vspace{-0.4cm}
\label{fig:slots}
\end{figure}

We conduct some finetune experiments in which we utilize the pre-trained generative diffusion model to train many iterations at a certain range of timesteps. Figure~\ref{fig:slots} shows the results that we trained a certain range of timesteps in many iterations. 
The results illustrate that training at the one range of timesteps is easy to destroy the efforts of training so many iterations at the other range of timesteps. 

% \subsection{Rethink Timesteps Sampler}

% And then, analyze the impacts. Next, we train the model several times at a different range of timesteps to see and analyze the effectiveness.

% We conduct some finetune experiments in which we first utilize the pre-trained generative diffusion model to train many iterations at a range of timesteps, individually. And then, analyze the impacts. Next, we train the model several times at a different range of timesteps to see and analyze the effectiveness. 
% Specifically, training the last range of timesteps in many iterations, and then train the first range of timesteps in several iterations. 
% The results illustrate that training at the one range of timesteps is easy to destroy the efforts of training so many iterations at the other range of timesteps. 
% Figure~\ref{fig:tsampler} (g) shows the results that we trained the first range of timesteps in many iterations, and then train the last range of timesteps in several iterations. The results illustrate the training at the last range of timesteps also can destroy the effectiveness of the training at the first range of timesteps.

By a series of experiments and analysis, we can conclude that alternately training at different scale levels of the MSE loss is extremely hard to converge. If we train a diffusion model using a small batch size with a uniform sampler, the random sampling of timesteps would destroy the previous efforts due to the alternate training in different scale losses. In conclusion, we should sample the timesteps so that the MSE losses of all selected timesteps contain different scales as much as possible. In this way, different scale losses should be decreased together. 

%  这就是为什么，小的batch size ，会很难收敛的原因。每一次迭代选择出来的timesteps 对应的 MSE loss 尺度相差太大，使得一直在破坏之前的努力。
% 然后我们提出来一个基于 batch size 来划分 slots 的方法。使得每一个 batch 都坐落在不同的 scale 上面。通过这种方法让模型能够收敛更快。不至于作用被抵消。
% Based on the above experiments and analysis, we propose a timesteps sampler that each batch is allocated into a range of timesteps. Firstly, we calculate the MSE losses for each timestep via a pre-trained model. And then, for each range of timesteps, the average cumulative sum of the MSE loss of all timesteps in this range should be equal, when our limited resource only handles with a small batch size.

\noindent \textbf{Rethink the contributions of each timestep.}
In this section, we will exploit the contributions of each timestep. The main question is ``Does each timestep contribute the same effectiveness?''

\begin{figure}[t!]
\centering
\includegraphics[width=1.0\textwidth]{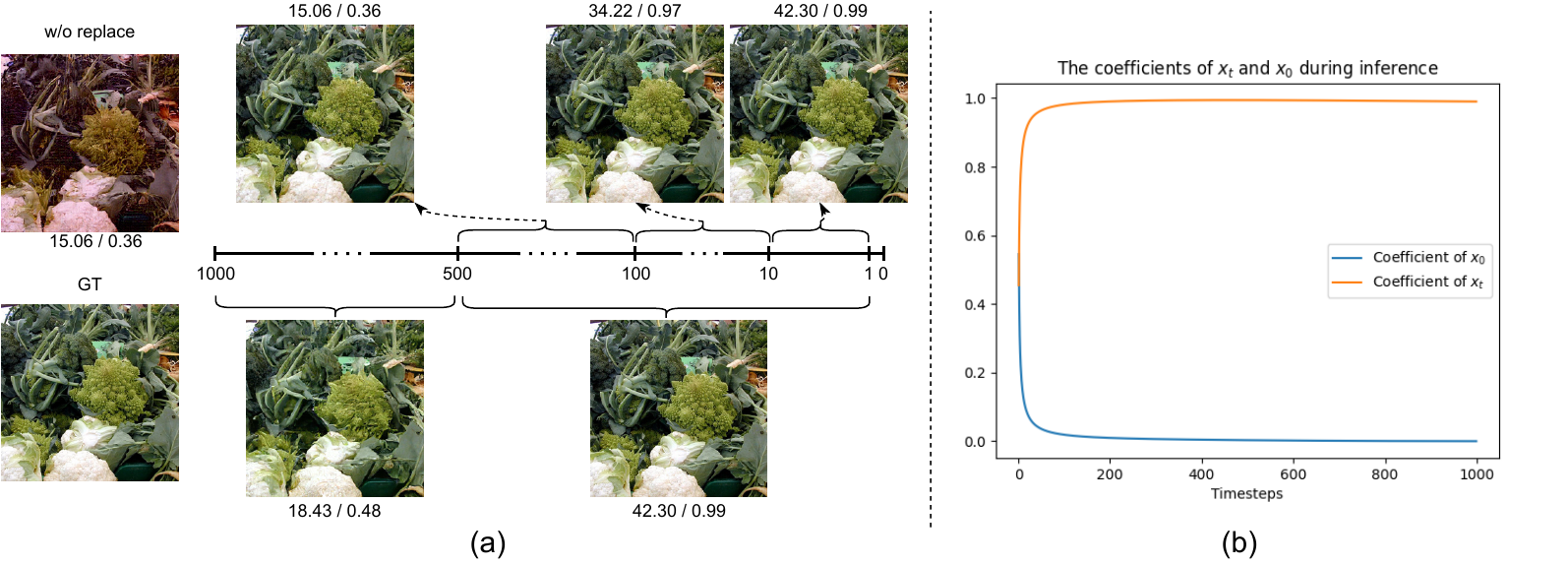}
      \caption{The contributions of each timestep.} 
% \textcolor{red}{[include more detailed descriptions.]}}
  \vspace{-0.4cm}
\label{fig:contributions}
\end{figure}

In order to exploit the contributions of each timestep, we focus on image super-resolution tasks via diffusion models. The diffusion model we used is the super-resolution diffusion model for ImageNet $64\times 64 \rightarrow  256\times 256$ from Guided-Diffusion~\cite{dhariwal2021diffusion}. We encountered the same issue that a small batch size produce a bad model with a uniform sampler. We infer some samples via the bad model. There are severe hue deviations in the generative pictures shown in the left top of Figure~\ref{fig:contributions}. The issue is very common in the generation of high-resolution images via diffusion models. There is a question: ``Which timesteps have more influence on the hue deviation issue?'' 

In the inference process from $t=T$ to $t=0$, we replace the predictions of $x_0$ via the real $x_0$ at a certain range of timesteps to see how much performance will be improved. Figure~\ref{fig:contributions} (a) illustrates the results of the experiments in that we replace a certain range of timesteps. As we can see, the performance of replacing the range of [1, 500] is better than replacing the range of [1000, 500]. Meanwhile, the performance of replacing the range of [10, 100] is better than replacing the range of [500, 100]. The performance is the best when we only replace the range of [1, 10] via the real $x_0$. The experiments prove that if a diffusion model can achieve well performance at the timesteps closer to 0, the better the performance. 
% the closer the timestep is to 0, the better the performance. 

Meanwhile, the coefficients of $x_t$ and $x_0$ of $p_{\theta}(x_{t-1}|x_{t}, x_0)$ during the inference process are shown in Figure~\ref{fig:contributions} (b). The coefficients of $x_0$ at most are small compared with the coefficients of $x_t$ from $t=T$ to $t=0$. Only at the timesteps nearly 0~(less than 15), do the coefficients of $x_0$ relative become larger. In the observations, we also prove the timesteps close to 0 contribute more. 
% \textcolor{red}{( + experiments)}

% \noindent \textbf{Conclusions}
% By the above experiments and analysis, we conclude that i) we should sample the timesteps that contain more different scale MSE losses between the prediction of noise $\boldsymbol{\epsilon}_\theta$ and the real noise $\boldsymbol{\epsilon}_t$ as much as possible at one iteration. In this way, different-scale MSE losses can decrease in all iterations so that different-scale MSE losses are decreased together and are not destroyed by each other. ii) The timesteps closer to 0 contribute more performance compared with the timesteps closer to $T$. We should focus more attention on the timesteps closer to 0. In other words, we should allocate more limited resources to the timesteps closer to 0 and scatter as much as possible to the timesteps with different-scale MSE losses. Therefore, we propose the timestep sampler. On limited resources, we have a relatively small batch size, denoted $B$. We split all timesteps $T$ into $B$ slots averagely based on the cumulative sum of the MSE losses between the prediction of noise $\boldsymbol{\epsilon}_\theta$ and the real noise $\boldsymbol{\epsilon}_t$ produced via the pre-trained model from $t=0$ to $t=T$. In this way, each batch will be responsible for one scale MSE loss for each iteration. Through many iterations, all scale MSE losses will be decreased together and not influence each other. 

\subsection{Rethink the prediction types of \texorpdfstring{$x_0$}.}
\noindent \textbf{Prediction Types.}
There are three parameterizations to predict $\mathbf{x}_0$ from the input $\mathbf{x}_t$ at time step $t$:
\begin{itemize}[leftmargin=*]
\item Predicting $\mathbf{x}_0$ directly, denoted \textit{d(istance)-prediction}.
\item Predicting $\mathbf{v}_t {=} \sqrt{\bar{\alpha}_t} \boldsymbol{\epsilon}_t {-} \sqrt{1 {-} \bar{\alpha}_t} \mathbf{x}_0$, which we can reparameterize the $\mathbf{x}_0$ via the input $\mathbf{x}_t$ and $\mathbf{v}_t$: $\mathbf{x}_0 {=} \sqrt{\bar{\alpha}_t} \mathbf{x}_t {-} \sqrt{1 {-} \bar{\alpha}_t} \mathbf{v}_t$. We denote it as \textit{v(elocity)-prediction} since we can obtain $\mathbf{v}_t {=} C \frac{\mathrm{d} \mathbf{x}_t }{\mathrm{d} t}$, where $C$ is a constant.
\item Predicting $\mathbf{\epsilon}_t$ from the input $\mathbf{x}_t$ at time step $t$: $\mathbf{x}_0 {=} \frac{1}{\sqrt{\bar{\alpha}_t}}(\mathbf{x}_t {-} \sqrt{1 {-} \bar{\alpha}_t}\boldsymbol{\epsilon}_t)$. We denote the original type of predictions for $\mathbf{x}_0$ as \textit{a(cceleration)-prediction} since we can obtain $\mathbf{\epsilon}_t {=} C \frac{\mathrm{d}^{2} \mathbf{x}_t }{\mathrm{d}^{2} t}$, where $C$ is a constant.
\end{itemize}
In empirical, a pre-trained model for the \textit{d-prediction} and the \textit{a-prediction} works better at time step $t$ closer to $t {=} T$ and $t {=} 0$, respectively. For the \textit{v-prediction}, it works more average for all time step $t$. To mitigate the bias of the prediction of $\mathbf{x}_0$ caused via different predicting types, we propose the mixed prediction of $\mathbf{x}_0$. 

\begin{figure}[t!]
\centering
\includegraphics[width=1.0\textwidth]{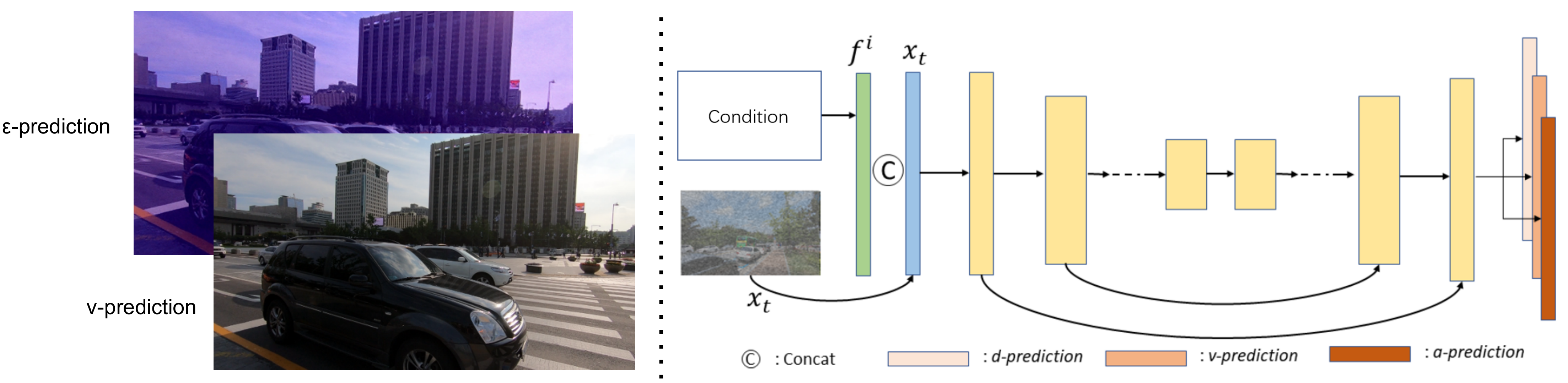}
      \caption{$v-$prediction in high-resolution generative diffusion models.} 
% \textcolor{red}{[include more detailed descriptions.]}}
  \vspace{-0.4cm}
\label{fig:v-prediction}
\end{figure}

\noindent \textbf{Rethink Prediction Types.} 
DDPM~\cite{ho2020denoising} and many of the following works choose to parameterize the denoising model by predicting $\epsilon_t$ with a neural network $\epsilon_\theta(x_t)$. There are common issues that the generated images have noise or hue deviations on high-resolution generative tasks. Especially, there exists the inconsistency of unnatural global color shifts across frames at video super-resolution tasks. But the frames from the $v$-prediction model do not and are more consistent. The $v$-prediction models are good at high-resolution super-resolution tasks, such as Figure~\ref{fig:v-prediction} shows the $v$-prediction can mitigate hue deviations in video super-resolution tasks that usually have high resolution. However, the performance of $\epsilon$-prediction models is better than $v$-prediction models in low-resolution generative tasks. For different tasks, the different prediction types of $x_0$ have different abilities. 

Meanwhile, in a certain diffusion model, do the different prediction types of $x_0$ have different abilities at each timestep? The Figure shows the different prediction types of $x_0$ have different abilities at each timestep.
% Beginning with the observation of super-resolution tasks via diffusion models, 

Therefore, we propose the Mixed-Predictions of $x_0$ shown in the right of Figure~\ref{fig:v-prediction}. We modified the last layer of the UNet model to generate three outputs for $x_0$, $v_t$, and $\epsilon_t$, respectively. In this way, we can let the model learn different prediction types of $x_0$ automatically. But there is an issue during the training process. We hope to select the minimal loss of the three predictions, but there are different values of the three predictions. A larger loss will affect more via gradient descent, which is opposite to our expectation. Usually, the three predictions differ by a large order of magnitude. Therefore, we modified the training strategy. During the training process, we select the minimal loss of the three predictions to do gradient descent. In this way, we can find the most accurate prediction type of $x_0$. However, the above ways can not result in a good approach to address the issue of training a good diffusion model with a small batch size. 

\section{Issues}
\subsection{Timesteps Sampler}
We have experimented with various timestep samplers, including approaches that allocate more weights for near $t=0$, and a sampler that decreases weight for near $t=0$ with increased weights for near $t=T$. Despite trying different designs for timestep sampling, it appears that we need to sample more timesteps for one batch if we want to obtain a good diffusion model. 
% There may be a good solution for how to design a timestep sampler. 
This raises the question: \textit{What is the optimal strategy for designing a timestep sampler that can effectively balance resource constraints while ensuring model performance?}

\subsection{Mixed Prediction of \texorpdfstring{$x_0$}.}
Achieving convergence with mixed predictions of $x_0$. has proven to be exceptionally challenging. In fact, the results obtained from mixed predictions are often worse than those from a single prediction of $x_0$, such as $\epsilon-prediction$. This leads to the question: Why is the mixed prediction of $x_0$ so difficult to optimize, and can a more effective method be devised to combine different prediction strategies for better performance?

I am currently grappling with these two issues and hope that researchers interested in overcoming these challenges will explore potential solutions, particularly in the context of training high-performing diffusion models with small batch sizes.

\bibliographystyle{splncs04}
\bibliography{miccai}

\begin{thebibliography}{1}
\providecommand{\url}[1]{\texttt{#1}}
\providecommand{\urlprefix}{URL }
\providecommand{\doi}[1]{https://doi.org/#1}

\bibitem{dhariwal2021diffusion}
Dhariwal, P., Nichol, A.: Diffusion models beat gans on image synthesis. Advances in Neural Information Processing Systems  \textbf{34} (2021)

\bibitem{ho2020denoising}
Ho, J., Jain, A., Abbeel, P.: Denoising diffusion probabilistic models. Advances in Neural Information Processing Systems  \textbf{33},  6840--6851 (2020)

\bibitem{nichol2021improved}
Nichol, A.Q., Dhariwal, P.: Improved denoising diffusion probabilistic models. In: International Conference on Machine Learning. pp. 8162--8171. PMLR (2021)

\bibitem{rombach2022high}
Rombach, R., Blattmann, A., Lorenz, D., Esser, P., Ommer, B.: High-resolution image synthesis with latent diffusion models. In: Proceedings of the IEEE/CVF Conference on Computer Vision and Pattern Recognition. pp. 10684--10695 (2022)

\bibitem{ronneberger2015u}
Ronneberger, O., Fischer, P., Brox, T.: U-net: Convolutional networks for biomedical image segmentation. In: International Conference on Medical image computing and computer-assisted intervention (2015)

\bibitem{sohl2015deep}
Sohl-Dickstein, J., Weiss, E., Maheswaranathan, N., Ganguli, S.: Deep unsupervised learning using nonequilibrium thermodynamics. In: International Conference on Machine Learning. pp. 2256--2265. PMLR (2015)

\bibitem{song2020denoising}
Song, J., Meng, C., Ermon, S.: Denoising diffusion implicit models. arXiv preprint arXiv:2010.02502  (2020)

\bibitem{song2020score}
Song, Y., Sohl-Dickstein, J., Kingma, D.P., Kumar, A., Ermon, S., Poole, B.: Score-based generative modeling through stochastic differential equations. arXiv preprint arXiv:2011.13456  (2020)

\end{thebibliography}

\end{document}